\documentstyle[palatino,11pt]{article}

\def\slide{}
\def\slideoff{}

\def\Sasa{Sa\v{s}a}
\def\Buvac{Buva\v{c}}
\def\Lucja{{\L}ucja}
\def\Iwanska{Iwa\'nska}
\def\Nunez{N\'u\~nez}
\def\simpliciter{\it simpliciter}

\begin{document}

\thispagestyle{empty}

\noindent
{~}\\[1.0in]
{\Huge\it\bf Context as a Spurious Concept} \\[60pt]
{\Large {\bf Graeme Hirst} \\[10pt]
Department of Computer Science \\
University of Toronto \\ 
Toronto, Ontario \\
Canada {\large M5S 3G4} \\[60pt]
Presented at the AAAI Fall Symposium on \\
Context in Knowledge Representation  \\ and Natural Language,  \\ 
Cambridge, Massachusetts, 8 November 1997} \\[2.0in]
{\small Copyright {\copyright} 1997 Graeme Hirst}

\newpage
\thispagestyle{empty}

\noindent
{~}\\[1.0in]
\begin{quote}
{\Large\bf Abstract}\\[10pt]
I take issue in this talk with AI formalizations of context, primarily
the formalization by McCarthy and {\Buvac}, that regard context as an
undefined primitive whose formalization can be the same in many
different kinds of AI tasks.  In particular, any theory of context in
natural language must take the special nature of natural language into
account and cannot regard context simply as an undefined primitive.
I show that there is no such thing as a coherent theory of context {\it
simpliciter}---context pure and simple---and that context in natural
language is not the same kind of thing as context in KR.  In natural
language, context is {\em constructed} by the speaker and the
interpreter, and both have considerable discretion in so doing.
Therefore, a formalization based on pre-defined contexts and
pre-defined `lifting axioms' cannot account for how context is used in
real-world language.
\end{quote}

\newpage
\setcounter{page}{1}

\section{Introduction}

I'd like to start with a generalization that I've made over the last few
years: 
\begin{quote}
The solution to any problem in AI may be found in the writings of
Wittgenstein, though the details of the imple\-men\-tation are sometimes
rather sketchy.
\end{quote}
Now, I'm not a scholar of Wittgenstein---in fact, the more I learn
about him, the less I feel I understand him---and I'm hardly going to
mention him in this talk.  Nonetheless, he's going to be there in the
background throughout the talk.  He'd have been a great choice as an
opening speaker for a symposium on context in KR and NL.  
\slideoff

My purpose in this talk will be to show that there is no such thing as
a coherent theory of context {\it simpliciter}---context pure and
simple; that context in natural language is not the same kind of thing
as context in KR; and that any theory of context in natural language
must take the special nature of natural language into account and
cannot regard context simply as an undefined primitive.

Before I proceed, I'd like to draw attention to the cogent paper by
Varol Akman, ``Context as a social construct'', which he'll be
presenting later today.  Akman says many things in his paper that I
wish {\em I'd} thought to say.  (I don't mean to imply, however, that I
agree with {\em everything} that Akman says, let alone that he agrees
with everything that I'm going to say in this talk).

\section{Context is as context does}

I want to begin with what might seem to be a completely different
topic---stressors.  Its relevance will become clear in due course.

Hans Selye (1950) coined the word {\it stressor} to refer to anything that
causes stress---a deviation or distortion of a system from its normal
state.%
  \footnote{Selye's intent was to disambiguate the word {\it stress},
   which, in casual usage, can refer both to a cause and to its effect;
   Selye reserved it for the latter and introduced the new word for the
   former (Selye 1950:~9).} 
Selye was mostly concerned with physiological responses to biogenic
stressors, but much of the subsequent research concentrated on
psychological stressors.  Whether some thing or some event {\it is} a
psychological stressor depends entirely on how it is interpreted by the
person experiencing it.  From one book on the topic:
  \begin{quote}
   A stimulus becomes a stressor by virtue of the fact that it has,
   indeed, engendered a stress response.~\ldots\ Psychosocial stressors
   become stressors by virtue of the cognitive interpretation, or meaning,
   assigned to the stressor.~\ldots\ For example, a traffic jam is a
   neutral event; it only becomes a stressor by virtue of the fact that
   the driver interprets the traffic jam as a threatening or otherwise
   undesirable situation.  If the driver would interpret the traffic jam
   as having some positive or desirable aspect to it, no stress response
   is likely to evolve.  (Everly 1989: 6--7)
  \end{quote}
Consequently, psychological stressors can be almost anything from
significant life events such as marriage and bereavement to a headache,
a bad day at the office, a loud noise, sex or lack of sex, the number
4, the number 666 ({\it cf\/} Noshpitz and Coddington 1990).  And the
concept of `stressor' now has even wider currency;
ecologists, for example, will speak of the stressors of an ecosystem:
  \begin{quote} \slide
   This study identifies~\ldots\ ecological stressors to boreal
   lakes~\ldots\ The primary stressors are alterations in lake levels
   due to dam construction and removal; nutrient loading from the town
   site and marina; the sports fisheries; and contaminant loading from
   local~\ldots\ and long-range~\ldots\ atmospheric sources.  (Evans
   1997)
  \end{quote} \slideoff
Thus, stressors are defined solely in terms of their effects---in fact,
in terms of their effects on any particular person or system.  What
acts as a stressor for one person or system might have no effect at all
on someone or something else.  Just about anything can, in principle,
be a stressor of something else.

What this means is that we cannot have any theory of the concept of
`stressor' {\simpliciter}.  There is no procedure by which we can
determine whether or not a particular entity is a stressor just by
looking at its properties or attributes.  All we can do is actually
apply the putative stressor to possible victims, and see if at least
one of them experiences stress.  Certainly, we can talk of `frequent
stressors' or `likely stressors' of various kinds of stressees, and we
can have theories of what kinds of objects these are and recognition
procedures for them.  But it's only when we particularize in just this
way that we can have such theories and procedures.

In other words, `stressor' is not a natural kind.  And neither is
`context'.  Like `stressor', `context' is a concept that is defined
solely in terms of effects in a given situation.

Just about anything can, in principle, be a context.  Whether something
actually {\em is} a context can be determined only by its effect (which
I'll describe a little while later).  And what is a context in one
particular case might not be a context in another case.  What this
means is that we can't have any theory of the nature of a context.
There is no procedure by which we can determine whether or not a
particular entity is a context just from looking at its properties or
attributes.  All we can do is apply the putative context to possible
victims, and see if at least one of them experiences a contextual
effect.

Consequently, any approach to `context' {\simpliciter} that tries
to or purports to reify it, formalize it, or just speak of different
views of it is inherently misguided.

McCarthy and {\Buvac} (1997), for example, explicitly decline to give
any definition of `context'---it's as undefined as an element of a
group, says McCarthy in another paper (1996).  But then they proceed to
stipulate---despite the lack of definition---that contexts can be
formalized as first-class objects, all of the same formal type, that
they're things that propositions can be true in, and that they're
things that can be entered and exited and nested.  McCarthy and
{\Buvac} seem to see contexts as {\em containers} of some
kind, at least metaphorically speaking.  But that's just an assumption
that they make, and it's an assumption that they make so deeply that
they never even refer to it explicitly.  My point in this talk is that
`context' simply doesn't permit this kind of approach.

Now, understand here that I am {\em not} opposing the general
enterprise of formalizing abstracta.  On the contrary, it's an
enterprise that I engage in myself---for example in Hirst 1995, I wrote
of ``differences as first-class objects''.  Rather, I {\it am}
suggesting that formalization, when appropriate, should be just about
the final step, rather than the first step, in the understanding of a
putative concept.  And in the case of `context', it's clear that we are
still in a {\em very} early stage of understanding.

McCarthy (1996) suggests that a mathematical logic of contexts would be
analogous to the mathematical theory of groups.  But he himself points
out that group theory arose from observations that the algebraic
properties of integers under addition were the same as those of the
rationals under multiplication, and the appropriate
abstractions could then be made.  In the case of context, however, we
don't even have the observations yet.  Sure, we can devise a nice new
formal logic, and even call it a logic of context rather than, say, a
logic of snibs or snecks or some gensymed word.  But if the logic is to
have something to say about what the English word {\it context} is
about, then a little more work is in order.  In his section entitled
``Desiderata for a mathematical logic of context'', even though he
explicitly mentions applications in natural language, McCarthy lists
nothing more than a few matters of formalization, and despite the
heading, no desiderata deriving from the rather obvious need that a
logic of context account for what context does in natural language, nor
even the desideratum of finding out what such desiderata might be.

Even while side-stepping any definition of `context', McCarthy and
{\Buvac} (1997), do say that contexts are ``rich'' objects.  Yet they
never open them up and look at their internal structure, preferring
instead to follow a path analogous to the development of group theory,
though they do seem to want their work to be genuinely useful in AI
applications.  This is all a bit like saying that a course on the
algebra of groups, rings, and fields is the only qualification that
anyone needs in order to become a professional accountant or
bookkeeper.

So what we need to do is think a little bit about what a context is and
what it does.  (That's why I like Varol Akman's paper.  Akman wants to
formalize context too, and he does so in conjunction with careful
thinking about what context is.)

\section{Informal notions of context}

One of the purposes of a symposium such as this one is to explore
pre-theoretical notions of context.  While many researchers in AI talk
about ``context'', or use representations that implicitly or explicitly
act as ``context'' in some sense, the notion of context is still
pre-theoretical.  And this symposium (and, even more so, its
predecessor at IJCAI-95) was conceived by {\Lucja} {\Iwanska} and
others in recognition of that.

As academic researchers, our natural response to the announcement of a
meeting like this is to give a paper with a short preamble on ``Here's
what context means to me, and hence should mean to everyone'', and
then, as if this were all beyond dispute, immediately proceed with a
formalization, representation, or algorithm.  Or even to dispense with
the preamble, as if one's own view were just presumed to be accepted
universally.

But here, I'd like to take things a little more slowly, and cover what
is often just the preamble.  For one thing, even to speak of
``pre-theoretical notions of context'' implies that a theory of context
{\simpliciter} will eventuate in due course, and, as I've just said, I
don't think that there can be any such thing.  So I'd like to speak of
``informal'', rather than ``pre-theoretical'', notions of context, and
then restrict the discussion in such a way as to make theorizing and
formalization possible.

To do this, I'll start by pointing to the name of the symposium itself: 
\slide
It's not
  {\it Context in AI}, nor 
  {\it Context in Computer Vision}, nor 
  {\it Context in Swedish Politics, 1894--1902}, 
nor is it just plain 
  {\it Context}.
Rather, it's 
  {\it Context in Knowledge Representation and Natural Language}.  

Now, this is a very ambiguous name.  First, both the terms {\it
knowledge representation} and {\it natural language} can denote objects
of study, the enterprise of studying those objects, and, metonymously,
both of these at once.  I think {\Lucja} probably intended the
metonymous reading.  Second, {\it and} in English can mean both
intersection and union, and union additionally admits a distributive
reading.  So, loosely characterizing KR and NL as sets of topics or
concerns, there are three main interpretations possible:
  \begin{quote} \slide
   1.~~Context in (KR $\cap$ NL) \\
   2.~~Context in (KR $\cup$ NL) \\
   3.~~(Context in KR) $\cup$ (Context in NL)
  \end{quote} 
Notice that the third interpretation isn't necessarily the same as the
second.  It doesn't go without saying that it's meaningful to speak at
all of a unified notion of context in KR and NL, as in the second
interpretation.  Maybe only the third reading is meaningful---that is,
context in KR and context in NL are two qualitatively different things,
and the title of this symposium is a kind of zeugma or pun.  It's like
having a symposium named {\it Stressors of People and Boreal Lakes},
and talking about what a bad day at the office has in common with
alterations in lake levels due to dam construction and removal.
\slideoff

And, indeed, the word {\it context} really covers quite a board
territory.  In their excellent survey article on the formalization of
context, Akman and Surav (1996) show that there are many different
kinds and uses of context even just within AI.  Sure, there are
similarities among these different kinds of context---that's why we use
the same name for each---but it doesn't follow that everything we say
about one kind will automatically be true of another kind.  One of our
jobs at this symposium is to try to sort out the different meanings,
and not just to presuppose that no such work is needed.

So what I'll talk about next is why I think that context in KR and
formal reasoning is not the same as context in natural language.  It
will follow from this that there can't be any useful theory or
formalization of context {\simpliciter}, because the behaviour of
each kind of context is different.  As I said before, all we have is the
contextual effect.  

But first, I need to make a distinction that I would have liked to have
fitted in a little earlier.  I want to distinguish between `context'
and `element of context'.  When I spoke earlier of ``the effects of a
context'', what I really wanted to say was ``the effects of one or more
{\it elements} of context''.  But I couldn't actually say that earlier,
because at that point I was still granting the idea of `context' as a
primitive.  But now that we realize that we need to look at the
internal structure of a context, we can talk about its individual
elements and about their effects, either singly or in concert.

\section{Context in KR}

Now, my research is primarily in natural language and computational
linguistics, and I don't feel qualified to comment on or criticize 
proposals for the formalization and use of context in formal knowledge
representation and reasoning.  Rather, I'll accept this research at
face value, and then contrast it with what is required with regard to
context in natural language, a topic that I {\em do} feel qualified to
talk about.

So I see no problem with the basic idea, from McCarthy (1987), of
making axioms context-dependent in order to be able to state them at
the most convenient or useful level of generality, nor with the
suggestions as to the advantages that might be gained from doing this
that are set out by Shoham (1991) and Akman and Surav (1996).  McCarthy
and {\Buvac}'s well-known example in which different databases make
different assumptions regarding the price of airplane components shows
the benefits of this approach.

But at the same time it shows the limitations.  The example involves
formal, propositional reasoning, and the notion of a proposition being
true or false in a context.  It assumes that the assumptions made by
the databases are static; and indeed, the exercise of writing {\em any}
context-dependent axiom assumes that the ``home'' context for the axiom
is, in effect, pre-defined; that contexts can be usefully related by
generalization and specialization; and that lifting axioms can be
pre-defined to relate truth in one context to truth in another.  That's
fine in a formal system, but it doesn't get us very far with language,
to which I'll now turn.

\section{Context in natural language}

When it comes to talking about context in natural language, there is
overwhelming consensus, I believe, on at least one point:  context is a
source of information that can be used (is used, should be used, may be
used, must be used) by a language processor to reduce (or completely
eliminate) ambiguity, vagueness, or underspecification in its
interpretations of the utterances that it processes.  That's one of the
effects of context.  It constrains interpretation.

And, in addition, context also affects both what the speaker intends to
say in the first place and how he or she goes about doing that.  But
there are really two kinds of context in this---zeugma again.  One is
the situation in which the speaker, as an agent, forms the intent to do
something, which in this case is to communicate some message.  This
context constrains the agent's intent.  The other is the context that
the speaker uses as a source of information in creating that message,
deciding exactly how it is to be expressed to the particular hearer.
This context constrains the form of the communication and its exact
content.

A point that immediately arises from this is that the use of context in
natural language communication is a psychological construct that is not
directly concerned with truth, but rather with interpretation and
belief, with the {\em generation of meaning}.  Propositional truth is
involved only insofar as the interpreter may use their beliefs about
what is and isn't true when they form an interpretation of some
utterance.  In natural language, a context is not something that
propositions are ``true in''.  It's something that interpretations are
formed in, or, more precisely, formed {\em with}.  An interpretation
can be more or less vague or ambiguous, and more or less in accord with
the speaker's intent, and if the interpretation is a
proposition---which it needn't be, of course---then it might indeed
have a truth value---though it needn't, of course.

A second point that arises is that the speaker has considerable
discretion in the selection or construction of the context that is used
in forming the utterance.  That is, the speaker can (under constraints
that I'll get to shortly) choose which potential elements of context to
attend to and use and which ones to ignore.  Likewise, the interpreter
has considerable discretion in the construction of the
interpretation---and, notwithstanding anything the speaker does, has
discretion even in the selection or construction of the context that is
used to create the interpretation.  That is, the interpreter can also
choose which potential elements of context to attend to and interpret
with and which ones to ignore.  In an ideal, cooperative conversation,
the speaker and hearer will {\em harmonize} their contexts ({\it cf\/}
Regoczei and Hirst 1990), {\em negotiating} what they deem to be
relevant; but they're under no special obligation to do so.

What I'd really like to do at this point is read you about twenty
pages on this topic from Donald N. Levine's fascinating book {\it The
flight from ambiguity} (1985).  I don't have time for that, so we'll
have to make do with a couple of quotations and my attempts to summarize
Levine's discussion.

Levine's main point is that an aversion to ambiguity in communication,
and hence to the kind of discretionary interpretation that I've just
described, is a modern Western phenomenon, ``unique'' in world history
(p.~21).  ``Most if not all of the literate civilizations have
considered the cultivation of ambiguous locution to be a wonderful
art'', Levine says (p.~21--22), and goes on to give many examples.
For instance: 
  \begin{quote}
   `The Somali language is sinuous' [says David
   Laitin].~\ldots\ Political arguments and diplomatic messages take
   the form of alliterative poems, mastery of which is a key to
   prestige or power.  These poems typically begin with long, vague,
   circumlocutory preludes, introducing the theme at hand, which is
   then couched in allegory.~\ldots\ `A poetic message can be
   deliberately misinterpreted by the receiver, without his appearing
   to be stupid.  [The receiver may] go into further allegory, circling
   round the issue in other ways, to prevent direct confrontation.'
   [Laitin 1977, p.~39]  (Levine 1985, p.~23--24)
  \end{quote}

This approach to communication reaches its zenith in Amharic.  Levine
again:
  \begin{quote}
   One is considered a master of spoken Amharic only when one's speech
   is leavened with ambiguous nuances as a matter of course.  Even
   among the other people of Ethiopia, the Amhara have been noted for
   extremes of symbolism and subtlety in their everyday
   talk.~\ldots\ Amharic conversation abounds with general, evasive
   remarks, like~\ldots\ {\it Se\d{t}agn!} (`Give me!') [in which] the
   speaker fails to specify what it is he wants.  When the speaker [is
   asked] about the object he desires, his response still may not
   reveal what is really on his mind; and if it does, his interlocutor
   will likely as not interpret that response as a disguise.  (pp.~25--26)
  \end{quote}
Levine goes on to describe the various so-called ``wax and gold''
formulas of Amharic poetry which have two levels of interpretation, a
veneer and a deeper meaning.  The deeper meaning might depend upon
symbolism or allusion in the surface meaning.  In the most difficult
form, called ``inside the olive'', finding the esoteric meaning
requires the interpreter to find a completely different context.
But this isn't just a matter of poetry.  Levine writes:
  \begin{quote}
   The ambiguity symbolized by the formula `wax and gold' colors the
   entire fabric of Amhara life.  It patterns the speech and outlook of
   every Amhara.  When he talks, his words carry {\it double entendre}
   as a matter of course; when he listens, he is ever on the lookout
   for latent meanings and hidden motives.  As an Ethiopian
   anthropologist once told me, wax and gold is far more than a poetic
   formula; it is the Amhara `way of life'.  (pp.~27--28)
  \end{quote}

In other words, both the speaker and the interpreter have some
discretion in {\em choosing} and {\em constructing} the context in
which the interpretation is to be built.  And while this discretion
might be greater, and more explicitly licensed, in Amharic or Somali
than in English, it's true in English too, even if our politicians
don't routinely speak in alliterative allegorical poetry.  We see it
every day in our ordinary conversations (Devlin and Rosenberg 1996,
p.~18), in advertising, political discourse, poetry, humor, allusion,
persuasion and deception, negotiation of meaning, and misunderstanding
and its repair (Hirst, McRoy, Heeman, Edmonds, and Horton 1994).

Here's a very simple example:  In his first presidential campaign, Ross
Perot said 
  \begin{quote}
   If the United States approves NAFTA, the giant sucking sound that we
   hear will be the sound of thousands of jobs and factories
   disappearing to Mexico. 
  \end{quote}
Perot's remark was widely reported, and was frequently alluded to by
other speakers.  Three years later, the Reverend Jesse Jackson could 
write, in an article on Martin Luther King's ``I have a dream'' speech,  
  \begin{quote}
   The `giant sucking sound' is not merely American jobs going to NAFTA
   and GATT cheap labor zones. The giant sucking sound is that as jobs
   and education diminish, our youth are being sucked into the jail
   industrial complex.\footnote{``32 years later: The dream
   unfulfilled.''  {\it JaxFax}, {\bf 3}(34), 8 August 1995. \\
   http://www.cais.com/pcedge/test/rb/fx50824.html}
  \end{quote}
Neither Perot nor NAFTA had been previously mentioned in the article,
but Jackson could allude to them anyway by this phrase, and {\em make}
them part of the context.  A quick Web search easily finds hundreds of
such allusions to this one phrase.  To fully understand the speaker's
intent, the interpreter has to recognize the allusion and adjust the
context of interpretation accordingly.

But of course, having discretion in constructing context does not mean
having complete freedom.  Obviously, considerable constraints arise
from the message that is to be communicated, the circumstances under
which the communication occurs, and the mechanisms of language itself.
Even in Somali, the interpreter has to take the utterance itself as a
given.  And any language imposes rules as to how anaphora, for example,
are to be interpreted with respect to the preceding text, and so any
preceding text is necessarily an element of the context.  Some aspects
of the situation are also obligatorily included.  For example, Deborah
Tannen (1990) has shown that in American English, there are classes of
sentences for which the gender of the speaker determines pragmatic
aspects of the intended univocal interpretation, and it's therefore an
element of the context that the interpreter ignores at his or her
peril.

So what can actually be used as an element of context in natural
language?  Many other people have already offered inventories or
taxonomies of the kinds of things that a speaker or interpreter must or
may include in a context, and so I don't want to spend a lot of time
here going into details.  I need only point out that it includes just
about anything in the circumstances of the utterance, and just about
anything in the participants' knowledge or prior or current experience
({\it cf\/} Empson 1953).  So, Sperber and Wilson (1986) have argued in
detail that a speaker or listener can use any fact or belief about the
world that they have as an element of context.  Ferrari (1997),
emphasizes the multimodal aspects of communication; he divides elements
of context into those of linguistic context, perceptual context,
intentional context, and encyclopedic knowledge, and he includes the
message itself, along with all the circumstances of its utterance, in
what he calls the ``communicative situation''.  Zarri (1995)
distinguishes between the {\it a~priori} ``internal~/ static'' context
and the ``external~/ dynamic'' context.  The former includes knowledge of
the language itself, such as the lexicon.  Akman (1997) reiterates
seven dimensions of context from the work of Wendell Harris, and I'll
leave it to him to tell you about that later today.  Manfred Pinkal 
(1985) summarizes it all rather well:
  \begin{quote} \slide
   Aside from the surrounding deictic coordinates, aside from the
   immediate linguistic cotext and accompanying gestural expressions at
   closer view, the following determinants can influence the
   attribution of sense:  the entire frame of interaction, the
   individual biographies of the participants, the physical
   environment, the social embedding, the cultural and historical
   background, and---in addition to all of these---facts and dates no
   matter how far removed in time and space.  Roughly speaking,
   `context' can be 
   \\ \hspace*{1em}{\it ---I'd rather say ``draw on''--- } \\
   the whole world in relation to an utterance act.
   (Pinkal 1985, p.~36)
  \end{quote} \slideoff

So the discretion exercised by a speaker or an interpreter in constructing
a context is, in effect, a determination of what, among all this, is
and isn't {\em relevant} to the utterance that is to be interpreted
(Sperber and Wilson 1986).  But this leads us to a terminological
difficulty.  For something to even be considered for possible relevance
seems to imply that, regardless of the actual decision, that thing {\em
is} an element of the context---otherwise, how could it come to be
considered?  There are {\em two} intuitive notions of `context' here.
The first, which I've tacitly been using up to this point, is the set
of things that are used to build the interpretation with (Sperber and
Wilson 1986, p.~15); and the other is that, {\em plus} the things that
{\em aren't} used but nonetheless had a potential to have been used.
For example, you might say that because the gender of the speaker is
{\em sometimes} a factor in interpretation, it is therefore {\em
always} an element of context, even if the interpreter doesn't always
choose to use it.  But then, by the same argument, you'd have to say
that Ross Perot's giant sucking sound is always in the context because
a speaker can always make an utterance that alludes to it.  And by a
similar argument, {\em everything} is in {\em all} contexts.  But
that's not a very helpful view.  A middle ground, and one that I lean
toward, is to say that context involves a notion of {\em attention} to
account for things that are at least considered for use in constructing
the utterance or interpretation; to decide that something is {\em not}
to be used in forming an interpretation is, in a sense, to use it in
forming that interpretation!

We see then that context is both a psychological construct and, as
Akman says, a social construct, and it's a social construct both in the
sense that it is a construct {\em of} society and in the sense that it
is constructed socially---in all our communication and social
interactions---and constructed dynamically.  It's not just a matter of
moving axioms between pre-defined contexts with pre-defined lifting
axioms.

Given all this, research on context in natural language starts to look
quite familiar.  In fact, much (or maybe most) research in natural
language in AI for the last 25 years and more can be seen simply as
attempts to characterize context ({\it cf\/} Sowa 1995).  Roger
Schank's scripts (Schank and Abelson 1977, Schank and Riesbeck 1981)
and Gary Hendrix's partitioned semantic nets (1975) in the 1970s; my
own marker passing in knowledge bases in the 1980s (Hirst 1987);
present-day statistical approaches based on lexical co-occurrence;
my own group's recent use of {\em lexical chains} as
``cheap'' context for tasks such as segmenting discourse, finding
real-word spelling errors, and automatically creating hypertext (Morris
and Hirst 1991, Hirst and St-Onge 1998, Green 1997)---these are all
really just attempts to provide or construct contexts with which
utterances can be interpreted.

These approaches have had varying degrees of success.  Some were simply
wrong---that is, they made observably false assumptions about the
nature of language.  For example, Schank's scripts assumed that
situations always uniquely pre-determine the word meanings and
inferences that are applicable in the situation.  McCarthy and
{\Buvac}'s approach seems to be, in effect, Schankian.  {\Buvac}
(1996), for example, chooses between two homonymous meanings of the
word {\it bank} in a logical form based on the sentence {\it Vanja is
getting money at a bank} by assuming that all other words in the
sentence are unambiguous and can be used to find the exact right axiom
in the commonsense context.  As far as I can tell, the example relies
crucially on the assumption of univocality of the words {\it get}, {\it
money} and {\it at}, and if {\it at} were changed to {\it from}, the
method would fail on the resulting polysemy or metonymy.

Other AI approaches to interpretation in context were perhaps a little
more correct in principle, but still made unrealistic assumptions about
language or impractical assumptions about the knowledge sources upon
which they were supposed to draw; my own work on using semantic
associations in a knowledge base as a context for disambiguation should
probably go in this category.  Nonetheless, this approach at least had
the merit that interpretation was incremental, including the
construction of context.  It did not assume that parsing, let alone the
building of a logical form, can occur prior to any consideration of
context or to processes of disambiguation and interpretation.

So there's a sense in which just about all research in AI on natural
language {\em is} research on context.  And as we now see, it's somewhat
different from context in KR.

There's one obvious objection to making this distinction between
context in NL and context in knowledge representation and reasoning.
Proper interpretation of natural language, we've been told for many
years, {\em requires} knowledge representation and reasoning.  So we'd
better have a single theory of context that covers them both.

I have two responses to this.  First, despite Sperber and Wilson
(1986), it's becoming clear that while the knowledge used in
interpreting natural language is broad, the reasoning is shallow.
Although we can't yet characterize it precisely, it seems to be pretty
much limited to reasoning about quite simple commonsense knowledge,
knowledge of kinds, of associations, of typical situations, and even
typical utterances.  We don't do arbitrary reasoning in
interpretation.  So we don't need a very general theory.

Second, we probably {\em do} do arbitrary reasoning on arbitrary
knowledge when we {\em assimilate} interpretations---when we build and
refine our mental models.  But there's no reason to think that that
necessarily requires the same representations or mechanisms as are used
in creating the interpretation.  On the contrary, it's now clear that
the mind uses many {\em different} kinds of representations and
mechanisms.  And, of course, to the extent that using natural language
and representing and reasoning about knowledge are both cognitive
activities, there's no reason to think that they are characterized by
{\em any} AI-style formalizations---and plenty of reasons to think that
they aren't, as Lakoff and {\Nunez} (1997, p.~22--23) have argued.

\section{Context as a spurious concept}

So far, I've argued that the notion of `context' can be defined only
in terms of its effects in a particular situation.  Just as a stressor
is anything that stresses, one way or another, in at least one
situation, context is something that constrains, one way or another, in
at least one situation.  In the case of natural language, many
different kinds of things can be elements of context.  Context in
natural language is {\em constructed}, in part, by the speaker and the
interpreter---it's not the same as context in KR.

In this light, `context' {\simpliciter} can be seen to come
dangerously close to being a spurious or incoherent concept in much the
same way that `absolute motion' is a spurious concept (Peacocke 1992).
In fact, there are quite a number of parallels between the two.  In
both cases, we have an intuition about the putative concept, and a very
robust intuition at that.  In our daily lives, we use what seems
to be the notion of absolute motion in our navigation and moving
around, and, as high-school science teachers know, it's not an easy
notion to break away from.  For obvious cognitive reasons, it's a
psychologically compelling idea.  We only need to look at the history
of science to see how reluctantly it was given up, even by highly
educated people.  Yet now, a hundred years after the Michaelson--Morley
experiment, it seems so obvious that `absolute motion' is an
incoherent or spurious concept that it's hard to imagine how people
ever thought otherwise.  

Likewise, we use what seems to be a general notion of context when we
build our interpretations of everything in our daily lives; but this
is nothing more than an illusion that arises from our inability to
examine our own mental processes of reasoning and language
interpretation.  Matthew Dryer (1997) has recently shown that the idea
of sentence topic in linguistics, which has long been thought to be an
intuitively well-founded concept, is in fact a chimera---what Dryer
calls a ``metalinguistic illusion''.  Dryer has shown that just because
a sentence is about something, it doesn't follow that there's any
constituent in the sentence that's what the sentence is about.  All
there is is discourse topic, even if the discourse is just a single
sentence.  `Context' {\simpliciter} might turn out to be like
this---seemingly intuitively well-founded, but revealed as a chimera
upon deeper analysis.

And both absolute motion and context {\simpliciter} are easy to
formalize.  Cartesian coordinates work quite nicely for the former in
simple everyday applications.  For the latter, McCarthy and {\Buvac}'s
(1997) formalization of context {\simpliciter} can, under certain
assumptions, find the price of airplane parts and disambiguate two
homonymous senses of the word {\it bank} ({\Buvac} 1996).  But however
useful they are in local human day-to-day navigation, Cartesian
coordinates are not a very useful formalization for what is now known
about the nature of space and time in theoretical physics.  And simple
formalizations of context {\simpliciter} might work on toy examples,
but there's no reason to expect them to apply to real-world natural
language.  On the contrary, a little analysis of what `context'
actually is suggests that they won't.

\section{Conclusion}

I think that AI in general is sometimes just a bit too impetuous in its
desire to formalize things, and it tries to turn things into systems or
logics without fully understanding them, as if simply by doing so they
would thereby come to be understood.  Sometimes this works; and
sometimes it just leads to meaningless, ungrounded formal
systems---Lakoff and {\Nunez} (1997) again.  To someone with a hammer,
every screw looks like a nail.  And topics that deal with language,
cognition, and acting in and interpreting the world get more than their
share of this bad treatment.

This seems to arise from a combination of overenthusiasm for Western
scientific method and a misunderstanding of the nature of language that
borders on fear.  In this view, language is a messy and highly
imperfect medium that is not to be trusted, but rather must either
be sidestepped entirely or be beaten into submission by means of logic
and formalism.  This is pretty explicit in the work of Frege 
and Bertrand Russell (1918, p.~205), for example.  Maybe that's why
Russell looked up to Wittgenstein.  Wittgenstein had the guts (and the
brains) to engage the difficult questions of language that Russell
avoided, and to find some frightening answers---that some concepts
can't be defined by necessary and sufficient conditions, for example.
That leads to my second observation about AI and Wittgenstein: 
\begin{quote}
All AI knows how to do is carry on as if Wittgen\-stein 
had never existed.
\end{quote}
Nor Heidegger and Gadamer; nor Donald Levine; nor Sperber and Wilson;
nor George Lakoff; nor Herb Clark; nor Harvey Sacks and Emanuel
Schegloff and Harold Garfinkel and Erving Goffman.  And I carry on that
way too, at times---but at least {\em I} feel guilty about it.
\slideoff

So in this talk, I've been rather negative and pessimistic in places,
and I don't want to close on that kind of a note.  After all, one thing
that the field of artificial intelligence has certainly succeeded in
over the years is expressions of unbounded optimism.  So I want to
close by emphasizing that we {\em do} have a good chance of getting a
handle on `context'---but we need to avoid premature, uninformed
formalization.  Situation theory (Devlin 1991) seems to me to be one
especially good candidate.  There is a strong intuitive relationship
between the ideas of `context' and `situation'; situation theory has
been under development for many years; and computational and linguistic
concerns have been there from the start (Barwise and Perry 1983).  It
is heartening to see books such as that of Devlin and Rosenberg (1996),
who apply situation theory to real language in use and who say in their
preface that their greatest intellectual debt is to Harvey Sacks.  So I
think that work on formalizing context that uses situation theory, such
as that by Akman and Surav (1996, 1997) and Ferrari (1997), is pointing
us in the right general direction.  There are also many other promising
approaches to context---I can't possibly mention all the names---and
I'm looking forward to hearing about some of them in this symposium.

\newpage

\small
\section*{Acknowledgements}

I am indebted to {\L}ucja Iwa\'nska for asking me to write this paper;
and to Stephen Regoczei, for enabling me to write it, by means of many
discussions over the years on a number of the issues herein.  I was
also helped by discussions with, and provocations from, Nadia Talent
and Chrysanne DiMarco.  I am grateful to my fellow Canadian taxpayers
for a research grant from the Natural Sciences and Engineering Research
Council.

\def\hang{\hangindent=1em \hangafter=1 \noindent}

\section*{References}

\hang
Akman, Varol (1997).  ``Context as a social construct.''  {\it Working
notes, AAAI Fall Symposium on Context in Knowledge Representation and
Natural Language}, Cambridge, MA, 1--6.

\hang
Akman, Varol and Surav, Mehmet (1996).  ``Steps toward formalizing
context.'' {\it AI Magazine}, {\bf 17}(3), Fall 1996, 55--72.

\hang
Akman, Varol and Surav, Mehmet (1997).  ``The use of situation theory
in context modeling.'' {\it Computational Intelligence}, {\bf 13}(3),
August 1997, 427--438.

\hang
Barwise, Jon and Perry, John (1983).  {\it Situations and attitudes.}
Cambridge, MA: The MIT Press.

\hang
{\Buvac}, {\Sasa} (1996).  ``Resolving lexical ambiguity using a formal
theory of context.''  In:  van Deemter, Kees and Peters, Stanley (1996).
{\it Semantic ambiguity and underspecification}, Stanford, CA: CSLI
Publications.  101--124.

\hang
Devlin, Keith (1991).  {\it Logic and information.}  Cambridge
University Press.

\hang
Devlin, Keith and Rosenberg, Duska (1996).  {\it Language at work:
Analyzing communication breakdown in the workplace to inform systems
design}.  Stanford: CSLI Publications.

\hang
Dryer, Matthew (1997).  ``The myth of sentence topic.'' Unpublished.

\hang
Empson, William (1953).  {\it Seven types of ambiguity}, third
edition.   London: Chatto and Windus.

\hang
Evans, Marlene S. (1997).  ``Preserving and protecting boreal park
lakes'', National Hydrology Research Institute, University of
Saskatchewan, 26 June 1997. \\
http://ecsask65.innovplace.saskatoon.sk.ca/ \\ pages/current/conserv/preser.html

\hang
Everly, George S. Jr (1989).  {\it A clinical guide to the treatment of
the human stress response.}  New York: Plenum Press.

\hang
Ferrari, Giacomo (1997).  ``Types of context and their role in
multimodal communication.''  {\it Computational Intelligence}, {\bf
13}(3), August 1997, 414--426.

\hang
Green, Stephen (1997).  ``Building hypertext links in newspaper
articles using semantic similarity.''  {\it Third Workshop on
Applications of Natural Language to Information Systems (NLDB~'97)},
Vancouver, June 1997, 178--190.

\hang
Hendrix, Gary G. (1975).  ``Expanding the utility of semantic nets
through partitioning.''  {\it Advance papers of the 4th International
Joint Conference on Artificial Intelligence}, 115--121.

\hang
Hirst, Graeme (1987).  {\it Semantic interpretation and the resolution
of ambiguity.}  Cambridge University Press.

\hang
Hirst, Graeme (1995). ``Near-synonymy and the structure of lexical
knowledge.'' {\it Working notes, AAAI Spring Symposium on
Representation and Acquisition of Lexical Knowledge: Polysemy,
Ambiguity, and Generativity}, Stanford University, March 1995, 51--56.

\hang
Hirst, Graeme and St-Onge, David (1998).  ``Lexical chains as
representations of context for the detection and correction of
malapropisms''.  In:  Fellbaum, Christiane (editor), {\it WordNet: An
electronic lexical database and some of its applications}, Cambridge,
MA: The MIT Press, 1998.

\hang
Hirst, Graeme; McRoy, Susan; Heeman, Peter; Edmonds, Philip; and Horton,
Diane (1994).  ``Repairing conversational misunderstandings and
non-understandings.'' {\it Speech communication}, {\bf 15}(3--4),
December 1994, 213--229.

\hang
Laitin, David D. (1977).  {\it Politics, language, and thought.}
The University of Chicago Press.

\hang
Lakoff, George and {\Nunez}, Rafael E. (1997).  ``The metaphorical
structure of mathematics: Sketching out cognitive foundations for a
mind-based mathematics.''  In:  English, Lyn D. (editor), {\it
Mathematical reasoning: Analogies, metaphors, and images}.  Mahwah, NJ:
Lawrence Erlbaum Associates.  21--89. 

\hang
Levine, Donald N. (1985).  {\it The flight from ambiguity: Essays in
social and cultural theory}.  The University of Chicago Press.

\hang
McCarthy, John (1987).  ``Generality in artificial intelligence.''
{\it Communications of the ACM}, {\bf 30}(12), 1030--1035.

\hang
McCarthy, John (1996).  ``A logical AI approach to context.''
Unpublished note, 6 February 1996.   
\\ http://www-formal.stanford.edu/jmc/ \\ logical.html

\hang
McCarthy, John and {\Buvac}, {\Sasa} (1997).  ``Formalizing context
(expanded notes).''  In: Aliseda, Atocha; van Glabbeek Rob; and
Westerst{\aa}hl, Dag (editors), {\it Computing Natural Language}.
Center for the Study of Language and Information, Stanford University.
Reprinted in {\it Working notes, AAAI Fall Symposium on Context in
Knowledge Representation and Natural Language}, Cambridge, MA, 99--136.

\hang
Morris, Jane and Hirst, Graeme (1991).  ``Lexical cohesion, the
thesaurus, and the structure of text.''  {\it Computational
linguistics}, {\bf 17}(1), March 1991, 21--48.

\hang
Noshpitz, Joseph D. and Coddington, R. Dean (1990).  {\it Stressors and
the adjustment disorders.}  New York: John Wiley \& Sons.

\hang
Peacocke, Christopher (1992).  {\it A study of concepts}.  The MIT
Press.

\hang
Pinkal, Manfred (1985).   ``Kontextabh\"angigkeit, Vagheit,
Mehrdeutigkeit.''  In: Schwarze, Christ\-oph and Wunderlich, Dieter
(editors), {\it Handbuch der Lexicologie}, K\"onigstein: Athen\"aum
Verlag.  The quotation used is translated and cited in: Quast\-hoff, Uta
M.  ``Context.''  In:  Asher, R.E. (editor) {\it Encyclopedia of
Language and Linguistics}, Pergamon Press, 1994, 730--737.

\hang
Regoczei, Stephen and Hirst, Graeme.  ``The meaning triangle as a tool
for the acquisition of 
abstract, conceptual knowledge.''  {\it
International journal of man--machine studies}, {\bf 33}(5), November
1990, 505--520.

\hang
Russell, Bertrand Arthur William (1918).  ``The philosophy of logical
atomism.'' In {\it The philosophy of logical atomism and other essays
1914--19}, edited by John G. Slater (The collected papers of Bertrand
Russell, volume~8), London: George Allen \& Unwin, 1986.  157--244.

\hang
Schank, Roger C. and Abelson, Robert P. (1977)  {\it Scripts,
plans, goals, and understanding.}  Hillsdale, NJ: Lawrence Erlbaum
Associates.

\hang
Schank, Roger C.  and Riesbeck, Christopher K. (1981) ``The theory
behind the programs: A theory of context.''  In: Schank, Roger C.  and
Riesbeck, Christopher K. (editors),  {\it Inside computer
understanding: Five programs plus miniatures}, Hillsdale, NJ: Lawrence
Erlbaum Associates.

\hang
Selye, Hans (1950).  {\it The physiology and pathology of exposure to
stress.} Montreal: Acta.

\hang
Shoham, Yoav (1991).  ``Varieties of context.'' In: Lifschitz, Vladimir
(editor), {\it Artificial intelligence and the mathematical theory of
computation: Papers in honor of John McCarthy}, Academic Press.
393--407.

\hang
Sowa, John F. (1995).  ``Syntax, semantics, and pragmatics of
contexts.''  {\it Working notes, Workshop on Context in Natural Language
Processing}, International Joint Conference on Artificial Intelligence,
Montreal, August 1995, 145--154.

\hang
Sperber, Dan and Wilson, Deirdre (1986).  {\it Relevance: Communication
and cognition}.  Harvard University Press.

\hang
Tannen, Deborah (1990).  {\it You just don't understand: Women and men
in conversation.}  New York: William Morrow and Company.

\hang
Zarri, Gian Piero (1995).  ``$\,$`Internal' and `external' knowledge
context, and their use for the interpretation of natural language.''
{\it Working notes, Workshop on Context in Natural Language
Processing}, International Joint Conference on Artificial Intelligence,
Montreal, August 1995, 180--188.

\end{document}